\pdfoutput=1

\documentclass[11pt]{article}
\usepackage[table]{xcolor}

\usepackage[preprint]{acl}

\usepackage{times}
\usepackage{latexsym}

\usepackage[T1]{fontenc}

\usepackage[utf8]{inputenc}

\usepackage{microtype}

\usepackage{inconsolata}

\usepackage{graphicx}
\usepackage{multirow}
\usepackage{booktabs}
\usepackage{graphicx}
\usepackage{array}
\usepackage{amsmath}
\usepackage{amssymb}
\usepackage{setspace}
\usepackage{amsfonts}
\usepackage{tabularx}
\usepackage{mathtools}
\usepackage{enumitem}
\usepackage{soul}
\usepackage{adjustbox}

\setlist[itemize]{itemsep=0pt}    
\setlist[enumerate]{itemsep=0pt}  
\setlist[description]{itemsep=0pt} 

\DeclareMathOperator{\E}{\mathbb{E}}

\usepackage{algorithm}
\usepackage{algpseudocode}

\algrenewcommand\algorithmicrequire{\textbf{Input:}}
\algrenewcommand\algorithmicensure{\textbf{Output:}}
\algrenewcommand\alglinenumber[1]{\footnotesize #1:}

\title{What is a protest anyway? Codebook conceptualization is still a first-order concern in LLM-era classification}

\author{Andrew Halterman \\
  Michigan State University \\
  \texttt{halterm3@msu.edu} \\\And
  Katherine A. Keith \\
  Williams College\\
  \texttt{kak5@williams.edu} \\}

\begin{document}

\maketitle
\begin{abstract}
 Generative large language models (LLMs) are now used extensively for text classification in computational social science (CSS).
 In this work, focus on the steps before and after LLM prompting---\emph{conceptualization} of concepts to be classified and using LLM predictions in downstream \emph{statistical inference}---which we argue have been overlooked in much of LLM-era CSS. We claim LLMs can tempt analysts to skip the conceptualization step, creating conceptualization errors that bias downstream estimates.
 Using simulations, we show that this conceptualization-induced bias cannot be corrected for solely by increasing LLM accuracy or post-hoc bias correction methods. We conclude by reminding CSS analysts that conceptualization is still a first-order concern in the LLM-era and provide concrete advice on how to pursue low-cost, unbiased, low-variance downstream estimates. 
\end{abstract}


\section{Introduction}\label{sec:intro}

Text classification---measuring rich, complex concepts from documents---is a fundamental task in computational social science (CSS). ``Traditional'' text classification using supervised models is well understood and widely used as a social science methodology \citep{d2014separating, wilkerson2017large, grimmer2022text, barbera2021automated}. However, approaches to text classification have fundamentally shifted in the era of generative large language models (LLMs).

Analysts can now provide an LLM with a prompt consisting of the task description, the target document, and instructions to generate the classification prediction. 
This ``zero-shot'' approach to classification has led to a proliferation of optimistic work that claims that one can replace human annotators with LLMs \citep[\emph{inter alia}]{gilardi2023chatgpt,ziems2023can,tornberg2024large}.

LLMs clearly reduce analyst effort and annotation costs, potentially broadening access to text-based CSS techniques. However, there are tradeoffs to this ``faster and cheaper'' approach.
Previous work has found the performance of zero-shot LLMs is limited for more ``complex'' CSS tasks \citep{thalken2023modeling,bamman2024classification, halterman2025codebook} and other work has emphasized the pitfalls of using \emph{only} LLM-generated labels in downstream statistical inference \citep{egami2023using,gligoric-etal-2025-unconfident,baumann2025large}.

Complementing this prior work, we highlight the important but often overlooked step prior to LLM prompting: creating \emph{codebooks} that define the semantic classes of interest.  Note, in this work, we only focus on \emph{positivist tasks}---those for which there exists a gold standard label for each document conditional on a fully specified codebook---and we do not consider subjective, interpretivist, or truly ambiguous tasks. 
We spend the remainder of this paper expanding on and providing evidence for the following three main claims.


\textbf{Claim 1: Annotation errors can be decomposed into conceptualization and operationalization errors.} Under our positivist assumption, annotation errors, including errors in LLM-generated labels, are either from \emph{conceptualization}---errors resulting from incomplete\footnote{See \S\ref{subsec:complete} for discussion of ``complete'' versus ``incomplete'' codebooks. \citet{fariss2020measurement} use the term ``translation validity'' for this concept.} class definitions in a codebook---or from \emph{operationalization}---errors during annotation from incorrectly applying class definitions. This decomposition draws on the social science literature on measurement \citep{adcock2001measurement, fariss2020measurement}. We define expert annotators as those who do not make operationalization errors and note that different conceptualizations can yield different gold standard labels for the same set of documents (see \S\ref{sec:discern-concept} for examples).

\textbf{Claim 2: LLMs can tempt analysts to skip conceptualization.}
In the pre-LLM era, text classification projects required human annotations. Although these annotations were costly, they were a useful forcing mechanism: human annotators require detailed coding instructions, forcing analysts to carefully define concepts in codebooks. LLMs, in contrast, can ``fail silently'': they can generate plausible labels, even in the absence of well-defined classes.
In \S\ref{sec:discern-concept}, we provide real-world examples of computational social scientists short-cutting the conceptualization step when using LLMs.

\textbf{Claim 3: Neither post-hoc bias correction methods nor increasing LLM accuracy can overcome conceptualization-induced bias.}
The \emph{science} aspect of computational social science implies that annotations are used in \emph{downstream statistical inference}, e.g., a correlation (regression) analysis. 
In recent years, methodologists have proposed unbiased estimators that combine (noisy) LLM-predicted labels with gold-standard human labels, such as PPI \citep{angelopoulos2023prediction}, DSL \citep{egami2023using}, and CDI \citep{gligoric-etal-2025-unconfident}. 
Yet, expert annotators---who by our assumption do not make operationalization errors---can still produce incorrect labels if provided with ``incomplete'' codebooks. These errors will result in biased downstream estimates even after applying post-hoc correction methods such as PPI or DSL. 
Likewise, conceptualization-induced bias cannot be corrected solely by more accurate LLM classifiers: increasing LLM accuracy (with an incomplete codebook as input) will decrease variance but will not address bias.
We provide simulation experiments to support this claim in \S\ref{sec:simulation}. 

Our three claims reinforce the \textbf{thesis} of this paper: in LLM-era text classification, conceptualization \emph{remains} a first-order concern, as it was in pre-LLM era CSS text analysis
\citep{grimmer2013text,nguyen2020we}.





\textbf{Running Example: Protests.}
To ground our arguments in a real-world application, we use a running example of \textsc{Protest} classification, a well-studied task in social science research---see Figure~\ref{fig:protest}.
For instance, consider a political scientist who is interested in the effects of exposure to protests (independent variable) on anti-incumbent voteshare in a semi-autocratic country (dependent variable) varying by geographical region in a single time period.
They hypothesize that large, public, peaceful protests lower the perceived costs for others to express opposition. 
To test their theory, they collect a news corpus and aim to obtain binary protest labels for each document.

\begin{figure*}[t]
\centering
\includegraphics[width=0.9\textwidth]{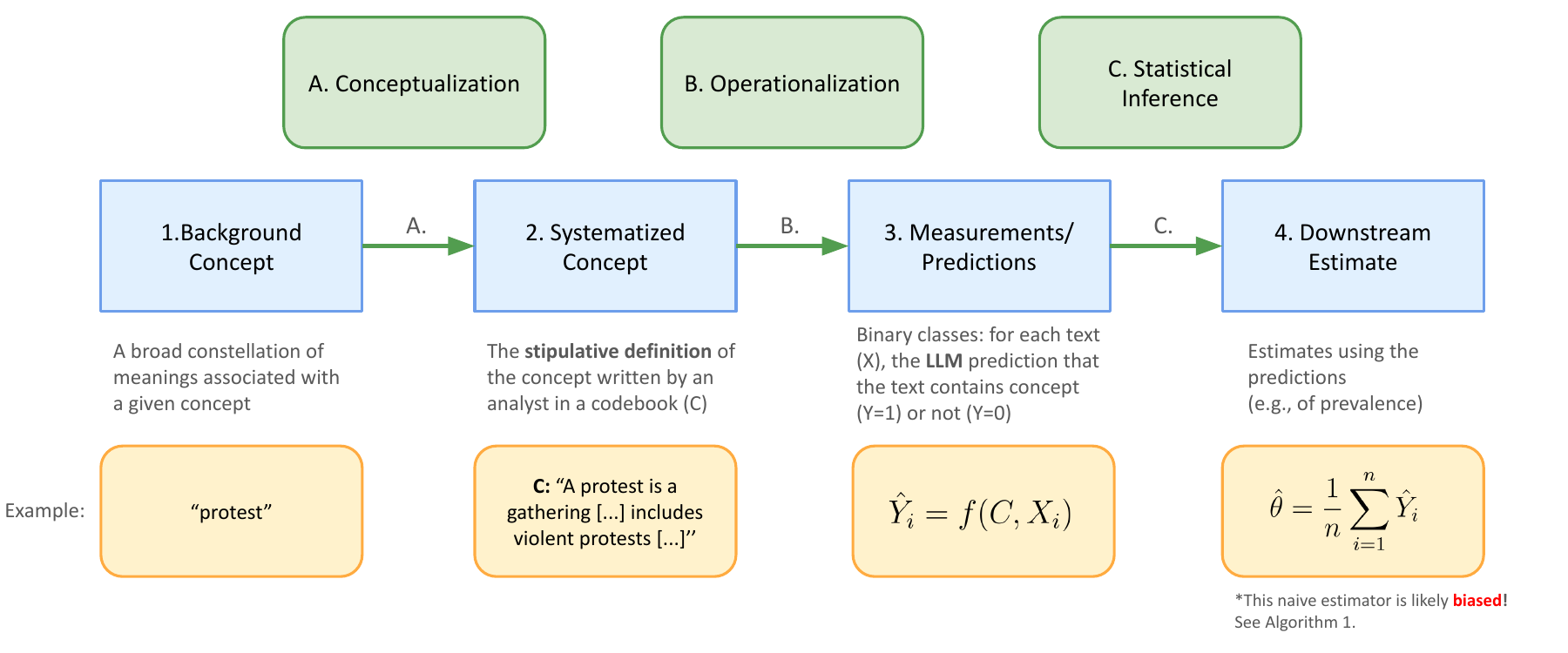}
\caption{\textbf{Text analysis with codebooks and LLMs}. \emph{Conceptualization} transforms a \emph{background concept} (e.g., ``a protest'') to a \emph{systematized concept} written in a codebook. In the current LLM-era, \emph{operationalization} consists of an LLM which takes as input the codebook and documents to make predictions. Predictions are then used in \emph{downstream estimates} of, e.g., the mean (prevalence) or correlation (regression) with other variables.
\label{fig:roadmap}
}
\end{figure*}

We incorporate this protest example into Figure~\ref{fig:roadmap} and also summarize the three main steps for classification in CSS: conceptualization, operationalization and downstream statistical inference.
In the next section, we illustrate possible approaches to our motivating protest example with a set of vignettes expanding Claims 1 and 2 (\S\ref{sec:vignette}).
We expand our conceptualization suggestions in \S\ref{sec:discern-concept} and operationalization suggestions in \S\ref{sec:discern-operation}.
Finally, in \S\ref{sec:conclusion}, we recommend a path forward for LLM-era CSS classification.

\section{Vignettes on Text Analysts}\label{sec:vignette}

To illustrate LLM-era text classification approaches, we use three stylized analyst vignettes.
We focus on a simple downstream inference goal: estimating the proportion of \textsc{Protest} documents in a corpus of size $N$: $\theta^* = \mathbb{E}[Y^*]$ where $Y^*=1$ indicates a news article contains a protest and $Y^*=0$ indicates an article does not. Each analyst seeks to obtain \textbf{unbiased} estimates of this parameter with \textbf{low variance} and \textbf{low cost} (in terms of time, compute and annotation expenses). We illustrate how an LLM pessimist, optimist, and pragmatist might approach this goal.

\subsection{Vignette: The LLM pessimist} 


First, imagine an ``LLM pessimist'' who believes \citet{baumann2025large}'s claim that ``100 human annotations outperform 100K LLM annotations'' and only use manual (human) annotations for their analysis. They follow a traditional, pre-LLM process: they spend approximately a year obtaining a grant,  then spend six months iteratively developing the codebook\footnote{We think these are realistic estimates as \citet{gerner2002conflict} note, ``CAMEO itself required about six months to develop, with between three and six people involved in
the process at various times.''}, and another six months obtaining the final gold-standard human annotations ($Y_j$) on $j=1, 2, \dots, n$ news articles where $n \ll N$. Then their point estimates and 95\% confidence intervals (CIs; under a normal approximation) are
\begin{align}
 \hat{\theta}_{\text{pess}} &= \frac{1}{n} \sum_{j=1}^n Y_j \\
   \textbf{95\% CIs:~} \hat{\theta}_{\text{pess}} &\pm 1.96\sqrt{\frac{\hat{\sigma}_y^2}{n}}
    \label{eqn:pess-ci}
\end{align}

\noindent
where $\hat{\sigma}^2_y$ is the empirical variance of the gold-standard annotations, $Y$. 

The advantage of the LLM pessimist approach is that their prevalence estimate, $\hat{\theta}_{\text{pess}}$, will be \textbf{unbiased} (assuming, of course, that their codebook is complete; an assumption we return to in \S\ref{sec:simulation}). 
However, this approach has the disadvantage that it is \textbf{extremely costly} in terms of analyst time and annotation expenses. 

\begin{algorithm*}
\caption{Prevalence (mean) estimator with LLMs; adapted from \citet{angelopoulos2023prediction}}
\begin{algorithmic}[1]
  \Require Unlabeled texts $\{{X_i} \}_{i=1}^N$, codebook $C$, texts with (gold-standard) human-labels $\{(X_j, Y_j)\}_{j=1}^n$, 
  LLM-predictions $\hat{Y} = f(\cdot)$, error level $\alpha=0.05$
  \State $\hat{\theta}^{f} \gets  \frac{1}{N}\sum_{i=1}^{N} f(C, X_i)$
  \Comment{LLM-only prevalence estimate}
  \State $\hat{\Delta} \gets \frac{1}{n}\sum_{j=1}^{n}\bigl(f(C, X_j)-Y_j\bigr)$
  \Comment{empirical ``rectifier'' using human labels}
  \State $\hat{\theta}^{\mathrm{PP}} \gets \hat{\theta}^{f}-\hat{\Delta}$
  \Comment{``prediction-powered'' estimate}
  \State $\hat{\sigma}^{2}_{f} \gets \frac{1}{N}\sum_{i=1}^{N}\!\left(f(C, X_i)-\hat{\theta}^{f}\right)^{2}$
  \Comment{empirical variance of LLM estimate}
  \State $\hat{\sigma}^{2}_{f-Y} \gets \frac{1}{n}\sum_{j=1}^{n}\!\left(f(C, X_j)-Y_j-\hat{\Delta}\right)^{2}$
  \Comment{empirical variance of empirical rectifier}
  \State $w \gets 1.96 \sqrt{\dfrac{\hat{\sigma}^{2}_{f}}{N} +\dfrac{\hat{\sigma}^{2}_{f-Y}}{n}}$
  \Comment{normal approximation to CIs}
  \Ensure Prevalence point estimate, $\hat{\theta}^{\mathrm{PP}}$, and 95\% confidence interval, $\mathcal{C}^{\mathrm{PP}}=
  \bigl(\hat{\theta}^{\mathrm{PP}}\pm w\bigr)$
\end{algorithmic}
\end{algorithm*}

\subsection{Vignette: The LLM optimist}

Most CSS work has presented an optimistic view of LLMs replacing annotators; notably, \citet{baumann2025large} systematically review 103 papers that use or benchmark LLMs in CSS and find ``88\% of reviewed papers recommend using LLMs for data annotation tasks.''
Let us imagine an individual political science PhD student (with a small budget) who trusts these optimistic recommendations. 
They input all $N$ news articles into an API-based LLM, $f(\cdot)$, with the prompt \emph{``Classify as protest: yes/no''}. Thus, their codebook definition ($\tilde{C}$) consists of only the surface-form of the class label, ``protest.'' They obtain a point estimate and 95\% CIs via 
\begin{align}
&\hat{\theta}_{\text{optimist}} = \frac{1}{N} \sum_{i=1}^N f(\tilde{C}, X_i) \\
    \textbf{95\% CI:~} &\hat{\theta}_{\text{optimist}}  \pm 1.96\sqrt{\frac{\hat{\sigma}^2_f}{N}}
\end{align}

\noindent
where $\hat{\sigma}^2_f$ is the empirical variance of the LLM predictions. 

The advantage of the optimist approach is that it is \textbf{extremely cheap}: the analyst spent minimal time conceptualizing the concept \textsc{Protest}, the LLM predictions from an API were far cheaper than hiring human annotators, and the analyst did not have to install software and use expensive hardware to train supervised models. However, it results in \textbf{biased} estimates ($E[\hat{\theta}_{\text{optimist}}] - \theta^* \not=0$) for two reasons. First, errors in the LLM's labels (operationalization error) lead to biased estimates \citep{angelopoulos2023prediction, egami2023using}. Second, their limited codebook likely results in a conceptualization error: $\E[Y | X, \tilde{C}] \neq \E[Y | X, C]$. For example, based on the downstream inference goal we described in \S\ref{sec:intro}, they should exclude \emph{violent} events from their definition of protests.  
We return to and expand this conceptualization concern in \S\ref{sec:discern-concept} and \S\ref{sec:simulation}.

\subsection{Vignette: The LLM pragmatist}\label{subsec:pragmatist}

Now, let us turn to a hybrid approach of an LLM pragmatist. The pragmatist analyst spends two weeks conceptualizing \textsc{Protest} and writing a bespoke definition of \textsc{Protest} in a codebook. They then spend one day annotating $n$ randomly sampled examples by hand. Finally, they spend $\$10$ inputting all $N$ news articles into an API-based LLM with their full codebook definition.\footnote{Alternatively, we describe in \S\ref{subsec:combine} how a pragmatist could also use older-generation supervised classifiers (possibly in combination with LLMs) and still arrive at unbiased estimates.} 
Combining their $n$ gold-standard annotations with the $N$ LLM predictions, they obtain a prevalence point estimate and confidence intervals via \citet{angelopoulos2023prediction}'s ``prediction-powered inference'' (PPI) algorithm, which we have adapted for codebooks and LLMs in Algorithm 1.

This pragmatist approach has advantage that the estimates are \textbf{unbiased}; \citeauthor{angelopoulos2023prediction} prove their estimator is unbiased and has valid coverage for any error level (assuming the human labels, $Y$, are equivalent to the true labels).
Also, the pragmatist approach is much \textbf{cheaper} than the labor-intensive approach of the pessimist. 
Because of these advantages, we recommend CSS analysts adopt this pragmatic approach in future studies. 
However, even a pragmatist can improve estimates by focusing on the following.  

\textbf{Avoiding conceptualization errors} that come from incomplete codebooks and result in statistical bias.  
In \S\ref{sec:discern-concept}, we provide real-world examples of possible definitions of \textsc{Protest} (that a pragmatist might use) and what constitutes conceptualization errors. 
In \S\ref{sec:simulation}, we show that even post-hoc bias correction methods cannot correct for conceptualization-induced bias.

    
\textbf{Reducing operationalization errors} from an LLM (increasing LLM accuracy) which reduces variance. 
    As we see in Line 6 of Algorithm 1, decreasing LLM operationalization (prediction) errors results in a decrease in $\hat{\sigma}^{2}_{f-Y}$ which results in estimates with lower variance (narrower CIs). 
    While improving LLM accuracy is a large focus of modern NLP research, we elaborate on particular concerns for CSS analysts in  \S\ref{sec:discern-operation}.

\begin{figure*}[t]
\begin{center}
    \includegraphics[width=0.92\textwidth]{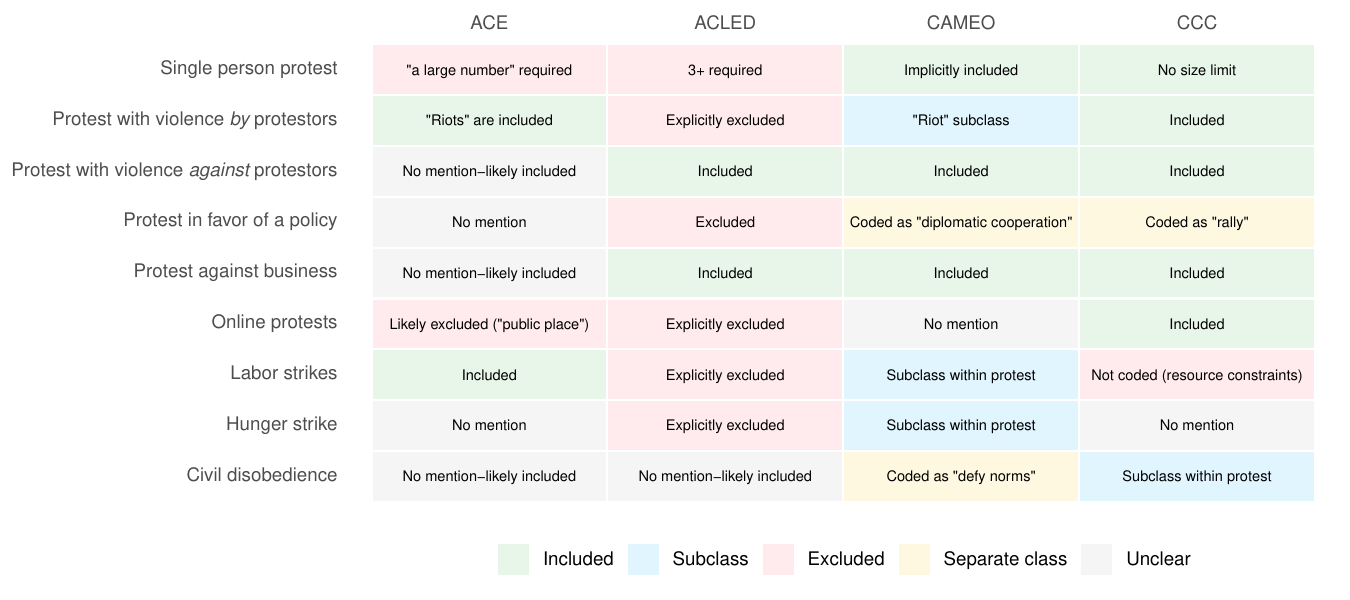}
\end{center}
\caption{
\textbf{Different stipulative definitions of \textsc{Protest} from real-world codebooks.} 
We manually categorize aspects of protest definitions from the codebooks of ACE \citep{doddington2004automatic}, ACLED \citep{raleigh2010introducing}, CAMEO \citep{gerner2002conflict}, and the \citet{ccc2024} (CCC). The length of the definitions also varies: from around 40 (white-space) tokens in ACLED, to around 100 for CCC and ACLED, to over 700 for CAMEO. See full definitions in \S\ref{app:protest-defn}.  
\label{fig:protest}}
\end{figure*}


\section{Pragmatists Aim to Avoid Codebook Conceptualization Errors} \label{sec:discern-concept}

\emph{Conceptualization} is the process of transforming a \emph{background concept} with a broad constellation of meanings (``protest'') into a \emph{systematized concept}---which we denote like \textsc{Protest}; see Step A in Figure~\ref{fig:roadmap}. In this section, we describe how concepts have traditionally been systematized in CSS, explain how LLMs can tempt analysts to skip the conceptualization step (Claim 2), provide guidance on when conceptualization is sufficient, and discuss tradeoffs with LLM-assisted codebook development.

\subsection{Definition types in codebooks can vary.}

In \citeauthor{adcock2001measurement}'s measurement framework, analysts transform background concepts into systematized concepts through \emph{definitions}, which we categorize into three levels of specificity:

\textbf{Type I.~Surface form of label.} The simplest definition consists solely of the surface form of the background concept, e.g., ``Is this a \emph{protest}?'' (the conceptualization used by the LLM optimist vignette). This assumes that the concept's meaning is self-evident and/or can be learned from background corpora.

\textbf{Type II.~Dictionary entry.} Alternatively, one could use a generic definition from a dictionary, e.g., ``A protest is a public (often organized) manifestation of dissent.''\footnote{WordNet \citep{miller1995wordnet} \url{https://en-word.net/lemma/protest}}
A dictionary definition is intended to apply to many downstream use-cases. 

\textbf{Type III.~Stipulative definition.} Most often in the social sciences, one needs a more specific definition than a dictionary entry. In this case, a domain expert could craft a \emph{stipulative definition}, a \emph{new} definition of a lexical unit for a specific context \citep{lycan1994stipulative,hitchcock2021definition}.
For example, \citet{raleigh2010introducing} defines \textsc{Protest} as \emph{``an in-person public demonstration of three or more participants in which the participants do not engage in violence, though violence may be used against them.''} This definition includes both inclusion criteria---a \textsc{Protest} must be in-person (not online) and in public---and an exclusion criterion---violence by protesters, and protests with only one or two participants are excluded.  

\subsection{The same background concept can map onto multiple systematized concepts}

For most substantive applications, stipulative definitions are required because the same background concept (e.g., ``protest'') can map onto multiple systematized concepts. Figure \ref{fig:protest} illustrates this variability by showing how four real-world codebooks define \textsc{Protest} with different stipulative definitions: the Automated Content Extraction (ACE) program \citep{doddington2004automatic}; the Armed Conflict Location and Event Data (ACLED), a widely used hand-coded dataset on violence and protest \citep{raleigh2010introducing}; the CAMEO event schema used by several machine-coded event datasets \citep{gerner2002conflict}, and the Crowd Counting Consortium (CCC)'s codebook for US-specific protests \citep{ccc2024}.

These codebooks differ as to whether a \textsc{Protest} includes  ``violence \emph{by} protesters'', the number of participants required, whether online protests are included, and several other aspects---see Appendix Section~\ref{app:protest-defn} for the full definitions. Thus, at inference time, the same text---e.g., the sentence ``a group of angry youth smashed the windows of businesses''---would result in different gold-standard class labels depending on the stipulative definition. We emphasize that this does not imply that one stipulative definition is ``more accurate'' than the others. Instead, they reflect different conceptualizations that are consequential for their specific downstream applications.

\subsection{Zero-shot LLMs can tempt analysts to skip conceptualization.}


The advent of zero-shot classification with LLMs has fundamentally altered the conceptualization process by removing the forcing function that expert annotators involved in the conceptualization process provided. In the pre-LLM era, expert annotators effectively required Type III stipulative definitions. Faced with ambiguous cases in early pilot rounds of annotation, annotators might ask clarifying questions such as: ``Do labor strikes count?'' ``How many people must be present?'' ``Does violence by protesters exclude an event from the \textsc{Protest} class?'' This forced analysts to develop precise stipulative definitions and explicitly address edge cases.

In contrast, LLMs can generate plausible labels even when provided only Type I or II definitions in their prompts. Instruction-tuned LLMs gained widespread appeal due to their high generalization\footnote{Focus on generalization with Type I definions has been a long trend in NLP including ``dataless classification'' \citep{chang2008importance} and applying general ``commonsense'' or ``world knowledge'' \citep[\emph{inter alia}]{zellers2018swag,sap2019social,bisk2020piqa}.} performance given simple natural language instructions (Type I definitions) such as ``Is the sentiment of this movie review positive or negative?'' \citep{weifinetuned}.  
This ability of LLMs to draw on concepts learned during training is useful, but creates the risk of ``silent failure'': the LLM can generate seemingly plausible labels without raising questions or signaling ambiguity an expert annotator would.


This temptation to skip conceptualization is evident in CSS papers in recent years in which researchers used zero-shot LLMs with Type I definitions without addressing conceptualization concerns.
For example,  \citet{halterman2021corpus}---who studied police inaction in the face of communal violence in India---used a zero-shot entailment model with a hypothesis ``Did police fail to act or not intervene?'' rather than a stipulative definition of ``intervention.''
\citet{brandt2024conflibert} provided LLMs with a document and list of event labels  without additional definitions. \citet{ziems2023can}'s experiments across 25 CSS benchmarks largely excluded class definitions from LLM prompts.

These current prompting practices---instructing the LLM to focus on generating particular labels---potentially limit the ability of the LLM to raise ambiguity. In the next section we discuss a complementary practice: how LLMs could possibly be used in initial pilot rounds of codebook conceptualization.

\subsection{Using LLMs for (semi-) automation of  codebook conceptualization has tradeoffs.}\label{sec:automated-codebook}

Recent work has proposed incorporating LLMs into the process of codebook conceptualization \citep{dai2023llm,Gao2023collab,xiong2025co,zhong2025hicode}. While these approaches have the advantages of decreasing analyst time used for conceptualization and potentially surfacing ``edge cases'', empirically evaluating their performance requires collecting expert labels across several versions of a codebook. Whether LLM-assisted codebooks overinflate analysts' confidence in the ``completeness'' of their codebooks is also an open question. 

\subsection{What constitutes a ``complete'' codebook?}\label{subsec:complete}

Although no codebook can address all possible edge cases, we propose two complementary standards for when a codebook is sufficiently complete.

\textbf{Community consensus on relevant aspects.} 
One completeness standard is that a codebook should address all aspects that a community of domain experts agree are relevant for defining class membership and possibly affect downstream inference.  For \textsc{Protest}, experts would likely judge a codebook as ``complete'' if it addresses all aspects shown in Figure~\ref{fig:protest}---participant violence, minimum size, location, violence \emph{against} protesters, etc.\footnote{As another applied example, consider the definition of civil war: how many annual battle deaths are required? Are anti-colonial wars included?} 
This echoes related considerations in causal inference around community consensus on treatment ``consistency'' \citep{hernan2016does} and the assumption of no unmeasured confounders.

\textbf{Expert agreement as an operational test.} Under our positivist assumption, a second practical standard is that two expert annotators, working independently with only the codebook (and no additional communication), should achieve near-perfect inter-annotator agreement, i.e. \emph{construct reliability} \citep{jacobs2021measurement}. 


These standards have an important implication: conceptualization cannot remain implicit. An analyst may have a clear mental model of \textsc{Protest}, but it must be written down in a  codebook\footnote{In the appendix we further formalize this claim; the ``reliability error'' row in Table~\ref{t:prag-four-type}.} to ensure \emph{test-retest reliability} \citep{jacobs2021measurement}, and then this complete codebook should be provided to both the expert annotators and LLM(s).  

\section{Pragmatists Aim to Reduce LLMs' Operationalization Errors}\label{sec:discern-operation}

Post-hoc correction methods guarantee unbiased estimates---as long as experts provide ``true'' labels under the positivist assumption---but analysts also want low-variance estimates. Increasing LLM accuracy (decreasing operationalization errors) yields downstream estimates with less variance (e.g., narrower CIs), but comes with cost tradeoffs. An enormous amount of work in academia and industry is focused on improving LLM accuracy. Here, we highlight a few points that we believe computational social scientists can actualize in their own work
---intervening on the codebook and prompt, the model, and/or the parsing of the output---and also highlight the cost tradeoffs.

\subsection{Optimizing prompts}

The easiest point of intervention for researchers seeking to improve LLM operationalization is to modify the inputs to the LLM, specifically the prompt format and the codebook. LLMs' accuracy is sensitive to even small changes in prompt format \citep{sclar2023quantifying, schulhoff2024prompt}, including in CSS \cite{atreja2025s}.

\textit{Longer codebooks increase computational costs.} The length and detail of codebooks imposes a tradeoff between conceptualization errors and operationalization errors. Adding longer clarifications, positive or negative examples, or ``FAQs'' on edge cases may more precisely define their classes, but LLMs are known to struggle with longer contexts, even prompts that fit within their maximum context length \citep{hsieh2024ruler,
levy2024same,liu2024lost,hong2025context}. 

\textit{Automated prompt optimization requires addition annotations.} Analysts may turn to automatic prompt engineering techniques such as AutoPrompt \citep{shin2020autoprompt}, DSPy \citep{khattab2023dspy}, or using LLMs themselves to optimize prompts \citep{zhou2022large}. However, these techniques still require (1) an initial prompt (DSPy calls this a ``signature'') and (2) gold-standard labeled examples against which the prompts are optimized.

\subsection{Optimizing models}

Second, researchers can reduce LLM operationalization error via the choice or modification of the LLM itself. 
In general, larger models are more accurate \citep{kaplan2020scaling} but require more compute (either locally or behind APIs). 
In addition to this obvious choice of \emph{which} LLM to use, applied CSS researchers may want to consider two other options that have been shown to increase performance, of course, with tradeoffs. 

\textit{Fine-tuning requires open-weight models and more annotations}. \citet{halterman2025codebook} show that parameter-efficient supervised fine-tuning (SFT) can improve LLM performance on three CSS datasets. However, SFT precludes the use of closed-weight LLMs and requires a large investment of time and computational resources. It also imposes additional data requirements: a gold standard example can be used either for SFT or for post-hoc bias correction, but not both.

\textit{Using multiple models increases computational costs}.\label{subsec:combine} An exciting area of current research is how to best combine multiple LLMs, or combine LLMs with older-generation (smaller or encoder-only) models.  \citet{pangakis2024knowledge} and \citet{halterman2023synthetically} propose using \emph{distillation} methods: using larger LLMs to generate ``silver standard'' labels (the teacher model) and then fine-tuning smaller open-weight models or supervised classifiers (the student models) on these generated labels.   

\subsection{Optimizing model outputs}

Analysts may also be able to improve performance by focusing on the outputs of an LLM (at inference time), again with costs. 

\textit{Scaling test-time compute increases computational costs}. Iterative inference methods, e.g., \emph{chain-of-thought} reasoning \citep{wei2022chain} and attempts to ``scale test-time compute'' \citep{snell2024scaling} have been shown to improve LLM accuracy, but at the cost of many more additional output tokens.

\textit{Calibrated probabilistic classifications require more annotations}.
 Prevalence estimation and other downstream inference estimators can often achieve better estimates by using ``soft'' classification probabilities, $\hat{P}(Y=1| X)$ \citep{keith2018uncertainty}. 
Obtaining well-calibrated probabilities from a generative LLM, either using token probabilities or ``verbalized confidence scores'' is still an open challenge \citep{tian2023just, xiong2024can}.\footnote{Even extracting hard classification labels from a generative model can be difficult \citep{schulhoff2024prompt}.} Future work could examine how to optimally allocate a portion of one's gold-standard annotation budget to supervised methods for calibrating confidence scores, i.e., Platt scaling \citep{platt1999probabilistic}.  

For a given CSS project, one (or multiple) technique(s) benefits of increasing LLM accuracy and  decreasing variance of downstream estimates may be worth the costs. However, we emphasize that none of these LLM techniques alone are able to correct for conceptualization-induced bias, which we elaborate on in the next section.







\begin{figure}
    \centering
    \includegraphics[width=\linewidth]{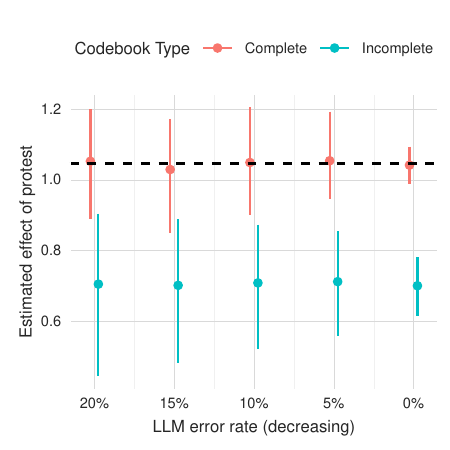}
    \caption{\textbf{ DSL-corrected estimates from simulated data} (\S\ref{sec:simulation}); we display the mean estimate (dot) and 95\% empirical intervals across $250$ simulations (bars). $N=10,000$ per simulation, and the true effect is the dashed line. \textbf{Takeaway:} Decreasing operationalization error reduces variance, but use of an incomplete codebook always results in biased estimates.  
    }
    \label{fig:sim-dsl}
\end{figure}

\section{Simulation Experiments} \label{sec:simulation}

Focusing on Claim 3, we use simulations to explore the effect of conceptualization errors (incomplete codebooks) on downstream inference. Here, we focus only on the pragmatist approach that combines LLM-generated labels and gold-standard (human) labels.\footnote{See \S\ref{app:simulation} for additional simulations under the ``pessimist'' and ``optimist'' paradigms.} 
To simplify experiments and avoid noise from text, we generate (simulated) structured data to represent different codebooks and annotations on documents. 
Building from our running protest example (Fig.~\ref{fig:protest}),
we simulate a \emph{complete} codebook that has a stipulative definition of \textsc{Protest} that excludes violence by the protesters. We simulate an \emph{incomplete} codebook as one that does \emph{not} address whether the stipulative definition of \textsc{Protest} includes or excludes violent events. 
We simulate a (true) linear DGP in which non-violent protests have a small positive effect on a dependent variable (the target estimand), and violent events have a strong negative effect on the dependent variable. 
We generate $n=1K$ gold-standard human annotations deterministically (given our positivist assumption) given the complete and incomplete codebooks, and generate $n=10K$ LLM annotations which have additional random noise added to the gold-standard labels. For estimation, we provide the gold-standard human annotations, LLM annotations, and dependent variable to the DSL correction method \citep{egami2023using}. We repeat the DGP under 250 different random seeds and use the empirical 95th percentile as the confidence interval. See \S\ref{app:simulation} for the full DGP and additional details.




The results in Figure~\ref{fig:sim-dsl} support Claim 3: the DSL correction yields unbiased estimates when the codebook is \textcolor[HTML]{F8766D}{complete}, with lower LLM errors yielding estimates with lower variance. In contrast, when the codebook is \textcolor[HTML]{00BFC4}{incomplete}, both the LLM and expert labels are systematically biased because the \emph{violence} aspect is not included in the codebook but affects the dependent variable in the regression. This reinforces our emphasis of ``complete'' codebook conceptualizations. 


\section{Conclusion and A Path Forward}\label{sec:conclusion}

LLMs can tempt CSS analysts to skip conceptualization, but incomplete codebooks induce bias that cannot be corrected by post-hoc methods or improved LLM accuracy. In CSS projects, LLMs are always used in a broader pipeline that involves careful definition of concepts and downstream statistical inference. We show that 
LLMs have already tempted analysts to speed through the conceptualization process and use the surface-form of labels (e.g., ``protest'') instead of full stipulative definitions. In the presence of a conceptualization error (e.g., not distinguishing a sub-aspect of a protest such as violence \emph{by} protesters), post-hoc bias correction methods will still be biased due to bias in the human annotations, posing a challenge for substantive conclusions for social sciences. These points reinforce our thesis: \textbf{conceptualization remains a first-order concern, even in the LLM-era}. 

Going forward, we make the following recommendations to CSS analysts who aim to obtain low-cost, unbiased, low-variance estimates. First, we recommend using the ``pragmatist'' approach which combines gold-standard annotations with LLM-generated predictions with post-hoc bias correction methods; although, it is an open problem as to which correction method performs best with finite samples for a given task, see \citet{de2025benchmarking}. 
Second, 
LLMs cannot fully replace the consensus from a community of peer experts on whether a codebook is ``complete'', so we recommend human domain expertise should be heavily incorporated into early pilot rounds of creating the codebook.
 Finally, analysts will likely be able to decrease variance in their estimates by moving beyond vanilla \emph{zero-shot} LLMs approaches, e.g., by increasing LLM test-time compute, fine-tuning an LLM on a supervised dataset, or using an ensemble of LLMs and supervised classifiers. Ultimately, we believe that thinking \emph{holistically} about the entire pipeline:  \emph{conceptualization}, \emph{operationalization}, and \emph{downstream statistical inference} can help solidify LLM-based classification as a rigorous methodology in the social sciences.

\clearpage
\newpage

\section*{Limitations}

Because our paper is primarily a conceptual one, we see only a few limitations to our work. First, as we describe in the introduction, we take a \emph{positivist} perspective and only focus on classification tasks that are not subjective. Post-hoc bias correction methods (e.g., DSL or PPI) also make this positivist assumption; however, there remain CSS text classification tasks that do not fall under this positivist paradigm. Second, for expository purposes, we only focus on a single running example from political science: protest classification. However, there are many other complex classification tasks from the social sciences that could similarly be examined using our framework. Finally, we note that the landscape of LLMs and NLP research is changing very quickly so there may rapidly emerge new LLM methods that help with conceptualization (see \S\ref{sec:automated-codebook}). 






\bibliography{main,latex/custom}

\appendix

\setcounter{figure}{0}
\setcounter{table}{0}
\renewcommand{\thefigure}{A\arabic{figure}}
\renewcommand{\thetable}{A\arabic{table}}

\section{Additional Related Work}

Our emphasis on \emph{conceptualization} connects to several other fields, which we summarize here.

\textbf{Lexical Semantics.} 
Linguists differentiate between semantic meaning derived from linguistic form and semantic meaning derived from the content to which speaker publicly commits \citep[p.~11]{bender2022linguistic}. A social scientist's \emph{act} of writing a new stipulative definition of a label in a codebook can be seen as a new public commitment to the meaning of that label. This act could be viewed as \emph{meaning transfer} or a \emph{deliberate semantic shift} \citep[p.~52]{bender2022linguistic} because there is typically a salient connection between a label's existing sense (the ``background concept'') and the ``nonce'' sense in the codebook. 

These deliberate semantic shifts are related to \emph{polysemy} (e.g.,~``kernel'' meaning something different in computer science, math, and nutrition) and the NLP tasks of \emph{semantic change detection} \citep{schlechtweg2020semeval,ehrenworth2023literary} and \emph{word sense induction} \citep{eyal2022large,lucy2023words}. However, codebook definitions differ in that they represent instantaneous, intentional meaning shifts, rather than natural semantic changes within a larger community over time.



\textbf{Data Benchmarks.} Echoes of our emphasis on conceptualization and local stipulative definitions also appear in work that formalizes and critiques data benchmarks for machine learning and NLP. \citet{raji2ai} warn against ``de-contextualized'' data and claims of ``general knowledge or general-purpose capabilities'' in data benchmarks. 
\citet{wallach2024evaluating} also observe that AI ``researchers and practitioners appear to jump straight from background concepts to measurement instruments, with little to no explicit systematization in between.'' 
\citet{zhou2025culture} call for the community to go beyond static cultural benchmarks and call for ``localization'' over ``generalization.'' \emph{Datasheets for Datasets} \citep{gebru2021datasheets} asks authors \emph{``Is there a label or target associated with each
instance? If so, please provide a description.''} However, they prescribe whether description should be Type II or Type III definition.   

\textbf{Other NLP work.} While our work focuses on corpus-centered classification (computational social science), other work aims to evaluate LLM generative output using LLMs themselves, the so called ``LLM-as-a-judge'' \citep{zheng2023judging}, or an ensemble/``jury'' of LLM judges \citep{verga2024replacing}.




\section{Full definitions of \textsc{Protest} from codebooks}\label{app:protest-defn}

This appendix provides the definitions of \textsc{Protest} used in the four codebooks we examine in the main text and Figure~\ref{fig:protest}.

\subsection{ACE}

ACE uses the class label ``demonstrate'' to refer to the concept of \textsc{Protest} in other datasets.

\begin{quote}
A DEMONSTRATE Event occurs whenever a large number of people come together in a public area to protest or demand some sort of official action. DEMONSTRATE Events include, but are not limited to, protests, sit-ins, strikes, and riots.
\end{quote}

\subsection{Crowd Counting Consortium (CCC)}

The CCC dataset classifies ``protests'' in a general sense. It includes a specific ``PROTEST" class, and distinguishes between \textsc{Protest}, \textsc{Rally}, \textsc{March}, etc. events. It defines the \textsc{Protest} class as

\begin{quote}
A crowd gathering in public to express disagreement with, or disapproval or anger or frustration toward, a specific individual or organization that is at or near the crowd’s gathering point (e.g., a politician giving a speech, a corporate headquarters, a bank branch, a construction site, a city hall), or in negative reaction to a recent or current event (e.g., the killing of George Floyd, the reversal of Roe v. Wade).
\end{quote}

\subsection{ACLED}

ACLED has the definition: 

\begin{quote}
A ‘Protests’ event is defined as an in-person public demonstration of three or more participants in which the participants do not engage in violence, though violence may be used against them. Events include individuals and groups who peacefully demonstrate against a political entity, government institution, policy, group, tradition, business, or other private institution. 
\end{quote}

In their codebook, they describe how the following are \emph{not} recorded as ‘Protests’ events: 
\begin{itemize}
    \item ``symbolic public acts such as displays of flags or public prayers (unless they are accompanied by a demonstration);''
    \item ``legislative protests, such as parliamentary walkouts or members of parliaments staying silent; 
    \item strikes (unless they are accompanied by a demonstration); and'' 
    \item ``individual acts such as self-harm actions like individual immolations or hunger strikes.''
\end{itemize}

\subsection{CAMEO}

The CAMEO codebook defines protests hierarchically. A protest is an event that fits any of the following subclasses:

\begin{itemize}
    \item Engage in political dissent, not otherwise specified
    \item Demonstrate or rally
    \item Conduct hunger strike
    \item Conduct strike or boycott
    \item Obstruct passage, block [in protest]
    \item Protest violently, riot
\end{itemize}

\begin{figure}[!ht]
    \centering
        \includegraphics[width=\columnwidth]{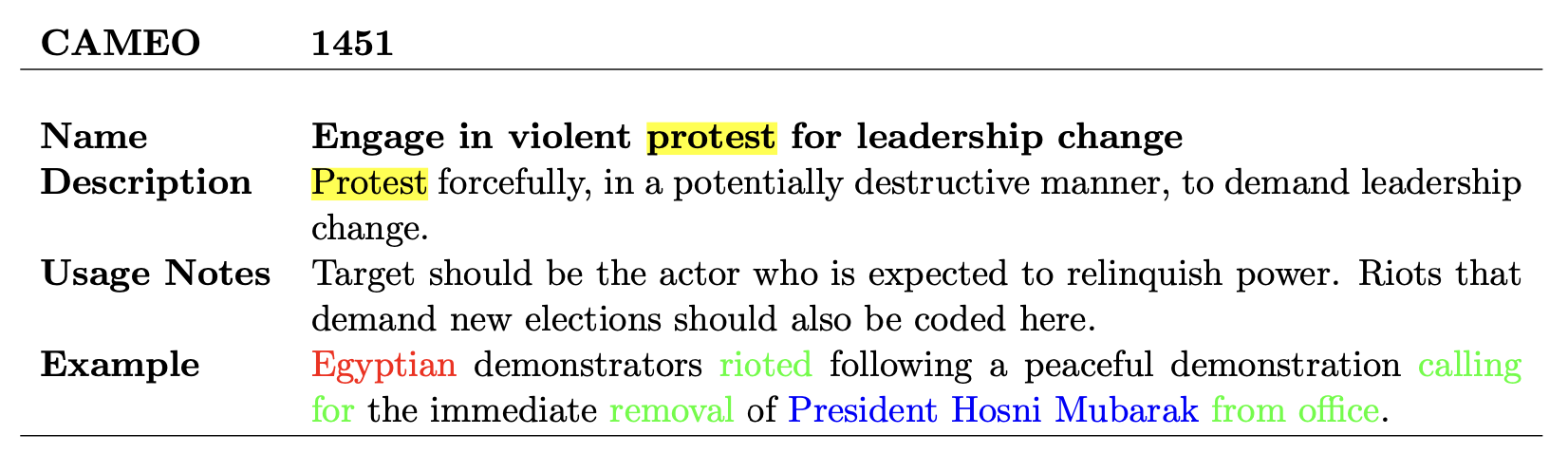}
    \caption{Example entry for the CAMEO codebook, illustrating the hierarchical definition of \textsc{Protest}. The ``protest'' class includes ``protest violently, riot'' as a subclass, which has a further subclass of ``engage in violent protest for leadership change''.}
    \label{fig:cameo}
\end{figure}

Each of these subclasses in turn has further subclasses referring to the target or demand of the protest. Each also includes examples, which we omit from the word count provided in the paper. Figure \ref{fig:cameo} illustrates an entry from the codebook.

\section{Breakdown of the pragmatic approach}

In Table~\ref{t:prag-four-type} we describe some additional ways one may approach downstream estimation goals. Our recommendation---the pragmatist approach with a complete codebook---is highlighted in green and the setting we explore in our simulation experiments (\S\ref{sec:simulation}) with an incomplete codebook is highlighted in red. 

We note there are two other settings which look similar to the pragmatist approach, but should be avoided. First there could be a \textbf{procedural error}, in which a complete codebook ($C$) is provided to the LLM but an incomplete codebook ($\tilde{C}$) is provided to the expert annotators. We note that if the complete codebook exists, it should also be provided to annotators. 
Second, there could be a \textbf{reliability} error in which an expert can provide gold-standard labels with the conceptualization ``in their own head'' and result in biased estimates. In \S\ref{subsec:complete}, we discuss how this violates \emph{construct reliability}.

\begin{table*}[t]
  \centering
  \resizebox{0.98\linewidth}{!}{ 
      \begin{tabular}{llll}
      \toprule
      LLM label & Gold label & Inference & Description \\
      \toprule
       -- & Y = Expert$(X, C)$ &Unbiased (high var.) & \textbf{Pessimist}. High variance from only human labels \\ 
       $\hat{\tilde{Y}}$ = LLM$(X, C)$ & 
     \hspace{1.1cm} --  & Biased & \textbf{Optimist.} No correction for operationalization errors \\
     \hline
       \rowcolor{green!10} $\hat{Y}$= LLM$(X, C)$ & Y = Expert$(X, C)$ & Unbiased & \textbf{Pragmatist}. A complete codebook permits post-hoc correction. \\ 
      $\hat{Y}$= LLM$(X, C)$ & $\tilde{Y}$ = Expert$(X, \tilde{C})$ & Biased & \textbf{Procedural error}. Gives incomplete $\tilde{C}$ to expert \\
      $\hat{\tilde{Y}}$ = LLM$(X, \tilde{C})$
      & Y = Expert$(X, \cdot)$ & Unbiased & \textbf{Reliability error}. Assumes an expert can label without $C$, see \S\ref{subsec:complete} \\
      \rowcolor{red!10} $\hat{\tilde{Y}}$ = LLM$(X, \tilde{C})$ & 
      $\tilde{Y}$ = Expert$(X, \tilde{C})$ & Biased & \textbf{Conceptualization error}. Incomplete codebook, see \S\ref{sec:simulation}\\
      \bottomrule
  \end{tabular}
  }
  \caption{
  \textbf{Comparing approaches to downstream inference}, many of which combine LLM zero-shot predictions with human expert gold-standard labels. Here, $C$ is a complete codebook and $\tilde{C}$ is an incomplete codebook. We
  \sethlcolor{green!10}\hl{highlight}
  the \textbf{pragmatist} row as the recommended approach, and also 
  \sethlcolor{red!10}\hl{highlight}
  the conceptualization error row which corresponds to Claim 3. \label{t:prag-four-type}}
\end{table*}

\section{Additional Simulation Details}  \label{app:simulation}

Here, we provide additional details on the simulation experiments described in \S\ref{sec:simulation}.

\begin{figure*}
    \centering
    \includegraphics[width=0.9\linewidth]{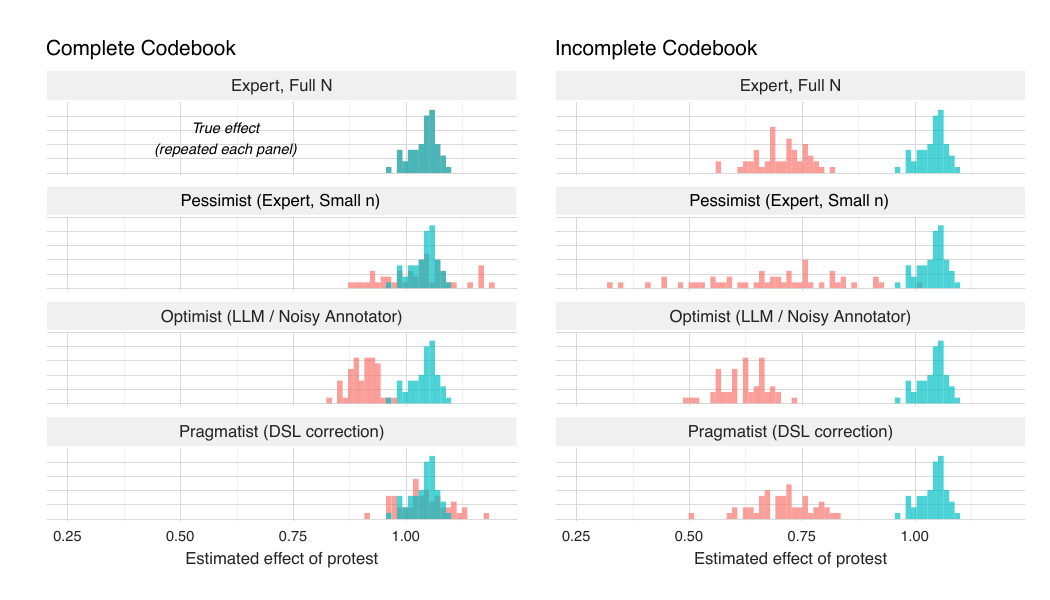}
    \caption{
    \textbf{Simulation results} comparing four estimation strategies across ``complete'' and ``incomplete'' codebook conditions across 50 simulations with $N = 10,000$ documents and a regression model where peaceful protests have a positive effect and violent protests have a negative effect. Results in \textcolor[HTML]{00BFC4}{blue} (repeated in each panel) show estimates from expert annotation of all $N$ documents. ``Small N'' and ``DSL'' conditions use 10\% expert annotated documents, and ``LLM'' uses labels for all $N$ documents with 10\% random error. The ``incomplete'' codebook omits the instruction to exclude violent protests.}
    \label{fig:simulation_results-app}
\end{figure*}

\subsection{True DGP}

Motivated by the real-world \textsc{Protest} codebooks (Figure~\ref{fig:protest}), we first simulate four (unimportant) aspects of a protest that will not individually affect the dependent variable. A political science domain expert (an author) selected the hyperparameters to represent a (fairly) realistic real-world scenario. For each instance, $i$, we sample  
\begin{itemize}
    \item  $Z^1_i \sim \text{Bernoulli}(0.2)$, meeting a basic protest definition
    \item $Z^2_i \sim \text{Bernoulli}(0.96)$, for events that are not hunger strikes, 
    \item $Z^3_i \sim \text{Bernoulli}(0.9)$, for protests \emph{against} someone
    \item $Z^4_i \sim \text{Bernoulli}(0.88)$, for greater than 3 attendees. 
\end{itemize}

Then, we sample whether the protest is violent (which is an important aspect, i.e., it affects the dependent variable directly):
\begin{equation}
V_i \sim \text{Bernoulli}(0.05)
\end{equation}

Using these aspects, we deterministically set the binary true protest label: 
\begin{equation}
D_i = Z^1_i \land Z^2_i \land Z^3_i \land Z^4_i \land \lnot V_i \label{eqn:d-def}
\end{equation}

We also sample a single \emph{covariate} that will also be input into the downstream estimate 
\begin{equation}
X_i \sim \mathcal{N}(0, 1)
\end{equation}
In a real-world example, this covariate might be something like voteshare in a prior election, the degree of urbanization in the geographic unit, or the union density of a region, each of which might be a confounder for the number of protests (independent variable) and later voteshare (dependent variable).

In our simulations, the dependent variable depends on whether a protest occurs or not \emph{and} whether or not the event is violent. This allows us to examine a realistic situation in which a codebook's completeness is correlated with the dependent variable. We simulate the dependent variable with the following regression equation:
\begin{equation}\label{sim:dgp}
Y_i = \beta_0 + \tau D_i + \beta_1 V_i + \beta_2 X_i + \epsilon_i
\end{equation}
\noindent
where 
\begin{equation}
\begin{aligned}
    \beta_0 &= -2 \\
    \tau &= 1 \\
    \beta_1 &= -5 \\
    \beta_2 &= 1 \\
    \epsilon_i &\sim \mathcal{N}(0, 1)
\end{aligned}
\end{equation}

Thus, the effect of true protests ($\tau$) is positive, but the effect of violent events ($\beta_1$) is negative.

\subsection{Simulating annotations}

Given the instances from the DGP, we simulate the following annotations. Following the pragmatist framework, we sample $n=1,000$ gold-standard (human) annotations and $N=10,000$ LLM annotations. Like the rest of our paper, our assumptions are that (1) gold-standard human annotations are under a \textbf{positivist} assumption that the expert human annotations will contain no operationalization errors, i.e., will completely adhere to the codebook they are provided; and (2) the LLMs are not perfectly accurate and will have a certain amount of operationalization error.  

\textbf{Complete codebook annotations.}
Under the assumptions above, we simulate the human labels under a complete codebook as a deterministic mapping from the true protest labels, 
\begin{equation}
A_i^{\text{complete}} = D_i. 
\end{equation}
Under the complete codebook, we sample LLM-generated labels as additive noise from the true protest labels with noise level $\delta$, 
\begin{align}
P_i &\sim \text{Uniform}(0,1) \\
\hat{A}_i^{\text{complete}} &= \begin{cases}
D_i &  \text{if} \hspace{1em} P_i > \delta \ \\
|D_i - 1| & \text{if} \hspace{1em} P_i \leq \delta \\
\end{cases}
\end{align}

\textbf{Incomplete codebook annotations.}
Under a \emph{complete codebook}, the true protest label ($D$) and the violent event label ($V$) are mutually exclusive, i.e., via Eqn.~\ref{eqn:d-def} we can never have an instance for which both $D_i=1$ and $V_i=1$. Under an \emph{incomplete} codebook, annotators (both human and LLM) only receive limited aspects of the protest definition, and they do \emph{not} have access to the aspect that excludes violent events. Thus, human annotations using an incomplete codebook are
\begin{equation}
A_i^{\text{incomplete}} = D_i \vee V_i.
\end{equation}

Likewise, the LLM is given the incomplete codebook so also does not know to exclude violent events and also annotates instances with error level $\delta$,
\begin{align}
P_i &\sim \text{Uniform}(0,1) \\
\hat{A}_i^{\text{incomplete}} &= \begin{cases}
D_i \vee V_i&  \text{if} \hspace{1em} P_i > \delta \ \\
|(D_i \vee V_i) - 1| & \text{if} \hspace{1em} P_i \leq \delta. \\
\end{cases}
\end{align}

\subsection{Estimation with DSL}
The inference goal is to estimate $\hat{\tau}$.
For each of the two codebook settings, as input into the DSL method \citep{egami2023using}, we provide the LLM labels $\{\hat{A}_i\}_{i=1}^N$, the covariates $\{X\}_{i=1}^N$, and the dependent variable $\{Y\}_{i=1}^N$ for all $N$ instances. We also input the gold-standard labels $\{A_i\}_{i=1}^n$ where $n \ll N$ (as we assume in \S\ref{sec:vignette}). Specifically, we fix $n = \frac{N}{10}$.
We also provide DSL with the assumed linear additive structure of the regression coefficients we aim to estimate, $Y \sim A + X$.

\subsection{Results}
As we show in Figure~\ref{fig:sim-dsl}, an incomplete codebook results in bias that cannot be corrected for, even as the LLM error approaches 0\%. 

In Figure \ref{fig:simulation_results-app}, we use the same DGP but also provide estimates for the other approaches (besides pragmatists) described in \S\ref{sec:vignette}. The regression setup here is similar to our straightforward example of mean estimation in \S\ref{sec:vignette}:  LLM ``optimists'' use the LLM-produced protest labels ($\hat{A}$) directly in a regression without any post-hoc bias correction. The LLM ``pessimists'' use only the gold-standard labels ($A$) in the regression. Similarly to the mean estimation in \S\ref{sec:vignette}, the ``optimist'' obtains biased regression coefficients, both from LLM operationalization error and conceptualization error, while the ``pessimist'' obtains unbiased estimates, but higher variance estimates from their smaller $n$.



\end{document}